# A method on selecting reliable samples based on fuzziness in positive and unlabeled learning

TingTing Li
College of Computer and Information Science,Chongqing Normal University
China
litingting@cqnu.edu.cn

WeiYa Fan
College of Computer and Information Science,Chongqing Normal University
China
fanweiya@cqnu.edu.cn

YunSong Luo
College of Computer and Information Science,Chongqing Normal University
China
luoyunsong@cqnu.edu.cn

## ABSTRACT

Traditional semi-supervised learning uses only labeled instances to train a classifier and then this classifier is utilized to classify unlabeled instances, while sometimes there are only positive instances which are elements of the target concept are available in the labeled set. Our research in this paper the design of learning algorithms from positive and unlabeled instances only. Among all the semi-supervised positive and unlabeled learning methods, it is a fundamental step to extract useful information from unlabeled instances. In this paper, we design a novel framework to take advantage of valid information in unlabeled instances. In essence, this framework mainly includes that (1) selects reliable negative instances through the fuzziness of the instances; (2) chooses new positive instances based on the fuzziness of the instances to expand the initial positive set, and we named these new instances as reliable positive instances; (3) uses data editing technique to filter out noise points with high fuzziness. The effectiveness of the presented algorithm is verified by comparative experiments on UCI dataset.

## 1 INTRODUCTION

Semi-supervised positive unlabeled (PU) learning refers to the task of learning a binary classifier from only positive and unlabeled data[1-3]. PU learning has drawn considerable attention recently, it is appealing to not only the academia but also the industry[4]. Such as:

Deceptive opinion detection[5], where the goal is to find negative deceptive opinions from the unlabeled ones.

Disease genes prediction[6], where the goal is to find disease candidate genes through gene expression profiles from the human genome.

Cross-domain sentiment classification[7], where the goal is to develop an instance adaptation method for cross-domain sentiment classification.

Software fault detection[8], where the goal is to build a prediction model for software fault detection.

In the traditional PU-learning, two methods are proposed to establish a PU classifier. One is to heuristically identify reliable negative (RN) instances from unlabeled instances and then use them to train standard binary classifiers, and the other is to use the whole unlabeled instances as negative data to establish binary classifiers[9].

The basic algorithm for PU-learning was described in text classification[10]. It proceeded as follows: first, the whole unlabeled set is considered as the negative class, the instances from the unlabeled set classified as negative are selected to form the initial set of RN. The second part iteratively enlarges the set of RN instances by increasing some additional instances from unlabeled set. The evaluation of the proposed method was carried out through used a set of hotel reviews gathered by Ott et al.[11]. As a further study, Donato et al[12]. believed that RN contains noise points, they proposed a method considers its iterative pruning, the experimental results demonstrated the effectiveness of the method.

Therefore, how to effectively select the reliable negative instances of unlabeled set is the most important work in PU learning. Some works proposed an automatic KL-divergence learning method which utilized the knowledge of unlabeled data distribution[9]. However, few people use the spatial structure of unlabeled set to extract valid information. In other words, we believe that unlabeled data may contain potential information of data space structure, which can be effectively used to assist PU-learning better and improve the generalization of PU-learning. As a knowledge mining tool, semi-supervised fuzzy c-means[13] can mine the data space structure information implied by a large number of unlabeled set. Yin et al[14]. proposed semi-supervised metric-based fuzzy clustering, which using the method of entropy regularization to solve the problem of parameter value selection of semi-supervised fuzzy c-means, and replacing the Euclidean distance with Mahalanobis distance, this way can select unlabeled instances with more information. The fuzziness of this classifier attains its minimum when every element absolutely belongs to the fuzzy set or absolutely not. The fuzziness attains its maximum when the membership degree of each element is equal to 0.5, i.e., the classification boundary is at the place where the fuzziness is equal to 0.5[15-17].

In this paper, we propose a novel framework to extract valid information of unlabeled instances. Our contributions are as follows:

1. In order to select reliable negative instances of the unlabeled set effectively, a new method that mines unlabeled instances space structure based on fuzziness is adopted. This is a general method which used on category data.

2. An approach to expand the initial positive set is proposed first, and we named these new instances as reliable positive instances.

3. With the help of classification fuzziness, we use data editing technique to filter out noise points.

A series of experiments are conducted on UCI dataset, using F-score as the measure. Experimental results show that our method can always reach high performance.

The rest of this paper is organized as follows: Section 2 presents related work; Section 3 introduces the new PU-learning framework based on semi-supervised metric-based fuzzy clustering algorithm; Finally, the experimental results are in Section 4; followed by the conclusion in Section 5.

## 2 Related work

### 2.1 PU Learning

PU-learning is a semi-supervised classifier with only positive instances (P) and unlabeled sets (U) that contain both positive and negative instances. Then our task is to construct a classifier (C), which can find all reliable negative instances (RN) hidden in the unlabeled set. It is formulated as follows: input $(P,U) \overset{C}{\Rightarrow}$ output(RN). The basic algorithm for PU-learning is shown in algorithm 1.

---

**Input:** a training dataset with a few positive samples P and lots of unlabeled samples U.
**Output:** the labeled dataset D
**Step 1:** i←1
**Step 2:** $C_i$←Generate_Classifier(P,U)
**Step 3:** $U_i^L$←$C_i$(U)
**Step 4:** $Q_i$←Extract_Negatives($U_i^L$)
**Step 5:** $RN_i$←$Q_i$
**Step 6:** $U_i$←U-$Q_i$
**Step 7:** while | $Q_i$|>Ø do
**Step 8:** i←i+1
**Step 9:** $C_i$←Generate_Classifier(P, $RN_{i-1}$)
**Step 10:** $U_i^L$←$C_i(U_{i-1})$
**Step 11:** $Q_i$←Extract_Negatives($U_i^L$)
**Step 12:** $U_i$←$U_{i-1}$-$Q_i$
**Step 13:** $RN_i$←$RN_{i-1}$+ $Q_i$
**Step 14:** Return($C_i$)

---

**Algorithm 1. The basic algorithm for PU-learning.**

### 2.2 Semi-supervised metric-based fuzzy clustering

In order to explore the effective information in unlabeled instances better, we hope to derive a group of clusters with side information and fuzzy nature of the data. Therefore, we need the classifiers whose outputs are vectors of membership grades. semi-supervised metric-based fuzzy clustering algorithm called SMUC is an algorithm that introduces metric learning and entropy regularization simultaneously into traditional fuzzy clustering. More specifically, SMUC focuses on learning a mahalanobis distance metric from side information given by the unlabeled instances. Thus, it makes the distance between instances within a cluster smaller than that between instances belonging to different clusters. Moreover, SMUC introduces maximum entropy as a regularized term in its objective function so that its formulas have more clear physical meaning compared with the other semi-supervised fuzzy c-means methods. The derivation formula of SMUC is as follows:

Given a prior membership degree and expressed as:

$$U' = \{u'_{ik} | u'_{ik} \in [0,1]; i = 1,...,n; k = 1,...,c\}, \quad (1)$$

where c and n are the numbers of clusters and instances, respectively. If $u'_{ik}$ is not given. Then the value $u'_{ik} = 0$, Clearly, conditions for $u'_{ik}$ can be written as:

$$\sum_{k=1}^{c} u'_{ik} \leq 1 . (\forall i = 1,2,...,n). \quad (2)$$

So, we formulate the SMUC algorithm as follows: Given a set of data instances $X = \{x_1, x_2, ..., x_n\}$ with some prior membership degrees $U'$, find the optimal membership U such that X can be properly partitioned c fuzzy subsets. First of all, we can obtain the preliminary centroid:

$$v'_k = \frac{\sum_{i=1}^{n} u'^2_{ik} x_i}{\sum_{i=1}^{n} u'^2_{ik}}. \quad (3)$$

Next, the covariance matrix is calculated as follows:

$$C = \frac{1}{n} \sum_{k=1}^{c} \sum_{i=1}^{n} u'^2_{ik}(x_i - v'_k)(x_i - v'_k)^T. \quad (4)$$

According to (4), the mahalanobis distance of x1 and x2 can be given as follows:

$$d_A^2(x_1,x_2) = (x_1 - x_2)^T C^{-1}(x_1 - x_2), \quad (5)$$

where $A = C^{-1}$. With this new metric, we minimize the dispersion between the instance and the set V of C prototypes:

$$J_A(U,V) = \sum_{k=1}^{c} \sum_{i=1}^{n} u_{ik} ||x_i - v_k||_A^2$$
$$+ \eta^{-1} \sum_{k=1}^{c} \sum_{i=1}^{n} (u_{ik} - u'_{ik}) \ln(u_{ik} - u'_{ik}), \quad (6)$$

where $u_{ik}$ is the membership degree of $x_i$ belonging to the cluster Ck whose centroid is $v'_k$ and satisfies the following condition:

$$s.t. \sum_{k=1}^{c} u_{ik} = 1, \quad u_{ik} \geq 0. \quad (7)$$

Equation (6) is the convexity of the optimization problem, we further define a Lagrangian function as follows:

$$L(U,V,\lambda) = J_A(U,V) + \sum_{i=1}^{n} \lambda_i \left( \sum_{k=1}^{c} u_{ik} - 1 \right)$$

$$= \sum_{k=1}^{c} \sum_{i=1}^{n} u_{ik} ||x_i - v_k|| + \eta \sum_{k=1}^{c} \sum_{i=1}^{n} (u_{ik} - u'_{ik}) \ln(u_{ik} - u'_{ik})$$

$$+ \sum_{i=1}^{n} \lambda_i \left( \sum_{k=1}^{c} u_{ik} - 1 \right), \tag{8}$$

where $\lambda_i$ is a lagrangian multiplier, and then:

$$\frac{\partial L}{\partial V_k} = \sum_{i=1}^{n} 2u_{ik}(x_i - V_k) = 0. \tag{9}$$

We get the optimal solution of $V_k$ as follows:

$$V_k = \frac{\sum_{i=1}^{n} u_{ik} x_i}{\sum_{i=1}^{n} u_{ik}}. \tag{10}$$

We take the derivative of $L(u_{ik})$ with respect to $u_{ik}$ and set it to zero:

$$\frac{\partial L}{\partial u_{ik}} = ||x_i - v_k||_A^2 + \eta^{-1}\left(\ln(u_{ik} - u'_{ik}) + 1\right) + \lambda_i$$

$$= 0. \tag{11}$$

Therefore

$$u_{ik} = u'_{ik} + \rho^{-\eta((d_{A(i,k)} + \lambda_i) + 1)}. \tag{12}$$

where $d_{A(i,k)} = ||x_i - v_k||_A^2$. According to constraints (7), we get the following equation:

$$\sum_{k=1}^{c} u_{ik} = \sum_{k=1}^{c} u'_{ik} + \sum_{k=1}^{c} \rho^{-\eta((d_{A(i,k)} + \lambda_i) + 1)}. \tag{13}$$

Reducing the above equation, we derive:

$$\rho^{-\eta \lambda_i} = \frac{1 - \sum_{k=1}^{c} u'_{ik}}{\rho^{-1} \sum_{k=1}^{c} \rho^{-\eta d_{A(i,k)}}}. \tag{14}$$

Then, the optimal solution of $u_{ik}$ as follows:

$$u_{ik} = u'_{ik} + \frac{\rho^{-\eta ||x_i - v_k||_A^2}}{\sum_{j=1}^{c} \rho^{-\eta ||x_i - v_j||_A^2}} \left(1 - \sum_{j=1}^{c} u'_{ik}\right). \tag{15}$$

Where each element in the output vector indicates the membership with which the input sample belongs to the corresponding class. The algorithm is given in Algorithm 2.

---

**Input:** D-dataset, a prior membership matrix $u'_{ik}$, number of clusters c.
**Output:** Membership matrix $u_{ik}$.
**Step 1:** Compute the preliminary centroid according to (3).
**Step 2:** Compute the metric matrix according to (4) using $u'_{ik}$.
**Step 3:** while $||u_k - u_{k-1}|| \geq \varepsilon$
  Compute $u_{ik}$ according to (15).
  Compute $V_k$ according to (10).
  end
**Step 4:** Return $u_{ik}$.

---

**Algorithm 2. The algorithm for SMUC. $u'_{ik}$, is the prior membership matrix, c is the number of clusters.**

## 2.3 Classification fuzziness and side effects of boundary instances

In [18], Zadeh first mentioned the term "fuzziness" and the concept of fuzzy set. This article points out the imprecision existing in ill-defined events, which cannot be described by sharply defined collection of points. Wang et al[15]. believed that the fuzziness of a fuzzy set attains its minimum when every element absolutely belongs to the fuzzy set or absolutely not, i.e., $u_{ik} = 0$ or $u_{ik} = 1$; they also find that instances with higher fuzziness are near to the classification boundary, while samples with lower fuzziness are relatively far from the classification boundary. The boundary is shown in Fig. 1 (a). Furthermore, it can be confirmed that both the misclassified instances and the instances with the larger fuzziness are all near to the classification boundary, which is shown in Fig. 1 (b).

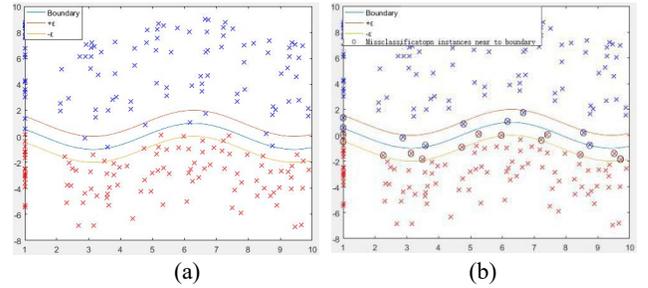

(a)                (b)

**Figure. 1. simple two class data and its boundary**

## 3 The main algorithm

Based on the analysis above, we propose a new PU-learning framework based on semi-supervised metric-based fuzzy clustering algorithm called PUFC. This algorithm calculates the cluster membership of U set bases on P set and U set, we select the membership degree that belongs to the positive class only. When the membership degree get closer to 1, the fuzziness of the data belongs to the positive class is smaller, that is, the probability of the data belonging to the positive class is larger. On the contrary, when the membership degree is closer to 1, the fuzziness of the data belonging to negative class is smaller, that is to say the possibility of being a negative class is larger. Therefore, we set a parameter $\varepsilon$ ($0 < \varepsilon < 0.5$), when the membership degree $u_i^+$ ($1 \leq u_i^+ \leq 0.5 - \varepsilon$), we consider this instance to be a reliable negative instance; when the membership degree $u_i^+$ ($0.5 + \varepsilon \leq u_i^+ \leq 1$), we consider this instance to be a reliable positive instance. In addition, the fuzziness and misclassification will reach its maximum when the membership degree of each instance is equivalent to 0.5. So, we consider instances with membership $u_i^+$ ($0.5 - \varepsilon < u_i^+ < 0.5 + \varepsilon$) as noisy points and edit these data. By means of this way it is possible to divide reliable negative instances, reliable positive instances and noisy points from the unlabeled set, and to construct a better final binary classifier than using the extra comparison algorithms. The specific algorithm is as follows:

---

**Input:** a training dataset with a few positive instances P and lots of unlabeled instances U, where $P = \{P_1, P_2, ... P_n\}$, $n$ is the number of positive instances, $U = \{U_1, U_2, ... U_m\}$, m is the number of unlabeled instances; Parameter ε.
**Output:** the labeled dataset D
**Step 1:** Set RN=Ø; Set RP=Ø; Set Noise=Ø.
**Step 2:** Set P with label 1, U with label -1.
**Step 3:** Clustering U set based on P and U with Algorithm 2, and obtaining cluster membership degree $u_{ik}$.
**Step 4:** Select the membership degree that belongs to the positive class $u_i^+$.
**Step 5:** for i=1:m
**Step 6:**  if $0 \leq u_i^+ \leq -\varepsilon + 0.5$
      RN=RN+$U_i$.
**Step 7:**  else if $-\varepsilon + 0.5 < u_i^+ < \varepsilon + 0.5$
      Noise= Noise +$U_i$.
**Step 8:**  else if $\varepsilon + 0.5 \leq u_i^+ \leq 1$
      RP= RP+$U_i$.
     end
    end
   end
  end
**Step 9:** The reliable negative instances RN is obtained, and set RN with label -1.
**Step 10:** The reliable positive instances RP is obtained, and set RP with label 1.
**Step 11:** Expand the initial positive set, where P=P+RP.
**Step 12:** Uses data editing technique to filter out noise point with high fuzziness, I.e. throw away the Noise set.
**Step 13:** Train a classifier C with P and RN.
**Step 14:** Classify instances in U using C.
**Step 15:** Output the labeled dataset D.

---

**Algorithm 3. The algorithm for PUFC**

## 4 Experiment and analysis

### 4.1 Experimental algorithm and description

In order to illustrate the effectiveness of the experiment, the following algorithm was used for the comparison experiment:
  (1) The basic algorithm of PU-learning（Basic）.
  (2) PU- learning based on spy technology (Spy)[19].
  (3) A conservative variant of the original PU learning algorithm. Instead of following an iterative growing strategy for building the RN set, this method considers its iterative pruning (Pruning)[12].
In our experiment, a collection of 9 real world dataset from the UCI and kaggle repository was used, where contain few hundreds/thousands of samples. Part of the dataset Ionosphere and dataset abalone are processed, and for dataset Ionosphere, two of the discrete attributes are deleted; For dataset abalone, we merge it into two categories of data. Table 1 shows some of their statistics.

**Table 1  Description of UCI dataset**

| Data | Size | Features |
|---|---|---|
| Banknote authentication | 1372 | 4 |
| Ionosphere | 351 | 32 |
| pima | 768 | 8 |
| Haberman Survival | 306 | 3 |
| Wholesale customers | 440 | 8 |
| Breast Cancer Coimbra | 115 | 7 |
| Breast Cancer Wisconsin | 516 | 30 |
| Indian Liver Patient | 583 | 5 |
| abalone | 4177 | 8 |

During the experiment, each dataset will be divided into two parts using ten cross-validation methods: the training set and the test set, in which the training set accounted for 90% and the test set accounted for 10%. Then 20%, 30%, 40%, 50% and 60% of the training instances were selected as the initial positive instances in the training set.

### 4.2 Evaluation measure

In our experiments, we use the popular F-measure on the positive class as the evaluation measure[20]. The F-measure is computed as follows:

$$F - \text{measure} = \frac{2 * \text{precision} * \text{recall}}{\text{precision} + \text{recall}} \quad (16)$$

The parameter ε is mentioned in the third part, we know that different parameter have different effects on the experimental results. Therefore, we assign different values to ε by taking 30% of the positive instances and calculate the F-measure of the algorithm proposed in this paper. Table 2 shows the effect on F-measure when selecting different parameter values.
It can be shown from Table 2 that different parameter values will produce different classification effects, and this phenomenon is related to the distribution characteristics of the dataset itself. Observe this table, we find that for different dataset, the parameters ε corresponding to the highest F-measure must be set in order to achieve the best experimental results. So in the next experiments, which 20% and 30% of the positive instances are selected as the initial positive instances in the training set. The experimental results are shown in Table 3 and Table 4, where the mean and standard deviation of the F-measure are provided in percent. The bold digit in each line indicates that the best mean F-measure obtains from the corresponding PUFC algorithm. correlation between attributes of dataset Wholesale customers or Indian Liver Patient are not strong. While SMUC takes into account the correlation between attributes in practical applications

**Table 2  The effect on F-measure when selecting different ε**

%(percent)

| Data | 0 | 0.05 | 0.1 | 0.15 | 0.2 | 0.25 | 0.3 | 0.35 | 0.4 | 0.45 |
|---|---|---|---|---|---|---|---|---|---|---|
| Banknote authentication | 72.2553 | **73.1030** | 72.4225 | 69.8368 | 67.6378 | 61.3922 | 61.4074 | 61.434 | 61.4903 | 61.4241 |
| Ionosphere | 77.4471 | 77.8932 | 78.1105 | 78.1806 | 77.7634 | 78.0532 | **79.7793** | 78.2814 | 77.673 | 77.8331 |
| pima | 80.7641 | 80.4503 | 79.8826 | 80.3955 | 80.6635 | **80.9097** | 80.615 | 79.9737 | 79.9684 | 78.8061 |
| Haberman Survival | 67.0818 | 65.9427 | 62.8308 | 65.1118 | 67.8659 | 70.0259 | 79.2072 | **84.6326** | 84.5817 | 84.5255 |
| Wholesale customers | 77.5154 | 79.6393 | 80.6274 | **80.7162** | 80.4627 | 80.5324 | 80.5975 | 80.6997 | 80.5711 | 80.6230 |
| Breast Cancer Coimbra | 61.4583 | 58.4975 | 55.3968 | 59.2451 | **63.5106** | 61.0724 | 61.2375 | 59.9638 | 59.2017 | 60.1415 |
| Breast Cancer Wisconsin | **89.5790** | 89.3284 | 88.6503 | 89.5432 | 82.8911 | 58.3785 | 55.5739 | 55.7431 | 55.4139 | 55.3528 |
| Indian Liver Patient | 44.7730 | 48.7804 | 57.0755 | 51.7715 | 57.2552 | 61.9700 | 60.0237 | 69.4509 | 70.6576 | **83.0750** |
| abalone | 49.8732 | 49.9642 | 50.1649 | 51.0339 | 50.7563 | 51.8679 | 52.3452 | 53.1425 | **53.6105** | 53.5071 |

**Table 3  F-measure corresponding to each dataset when the labeled instances set is 20%**

%(percent)

| Data | PUFC | Spy | Basic | Pruning |
|---|---|---|---|---|
| Banknote authentication | **73.3377±4.3797** | 42.8554±4.5199 | 44.5258±7.3272 | 45.1275±7.9486 |
| Ionosphere | **77.5766±7.4217** | 77.0215±6.9856 | 77.0226±8.8019 | 77.5504±5.0861 |
| pima | **80.5316±3.8007** | 67.8393±9.6371 | 74.8178±5.6586 | 75.5283±6.2065 |
| Haberman Survival | **84.5773±4.7251** | 40.4222±8.9044 | 42.7199±4.8383 | 42.1104±10.5274 |
| Wholesale customers | **80.5162±5.7551** | 80.4106±6.6747 | 80.2920±8.1469 | 80.3960±6.9866 |
| Breast Cancer Coimbra | **60.8730±18.7243** | 56.1534±19.1862 | 56.8126±17.2072 | 60.1249±15.1223 |
| Breast Cancer Wisconsin | **87.4444±4.7496** | 87.1651±5.4424 | 86.8066±8.5495 | 87.0548±6.0685 |
| Indian Liver Patient | **83.1898±3.7159** | 83.1542±4.0601 | 83.1342±4.2791 | 83.1061±4.5003 |
| abalone | **53.5474±1.7763** | 10.1553±2.7646 | 12.2736±4.5039 | 17.4083±5.0379 |

**Table 4  F-measure corresponding to each dataset when the labeled instances set is 30%**

%(percent)

| Data | PUFC | Spy | Basic | Pruning |
|---|---|---|---|---|
| Banknote authentication | **73.2512±5.9918** | 43.9693±5.8424 | 46.2228±7.5175 | 47.2304±6.7198 |
| Ionosphere | **79.8883±5.7919** | 77.4390±8.2798 | 77.8232±8.4845 | 78.4424±6.3602 |
| pima | **80.6545±3.5522** | 76.5370±4.8736 | 78.5755±3.1937 | 79.3060±4.0949 |
| Haberman Survival | **84.5646±4.8638** | 47.4618±12.0583 | 50.4281±8.7767 | 50.4026±10.1718 |
| Wholesale customers | **80.7162±2.3700** | 80.5005±5.6913 | 80.4568±6.3885 | 80.5479±5.2578 |
| Breast Cancer Coimbra | **62.4385±9.5240** | 60.0509±13.3866 | 61.2411±11.3887 | 61.3631±15.3276 |
| Breast Cancer Wisconsin | **89.6980±5.3182** | 88.2327±5.5703 | 88.9063±5.7307 | 88.8687±5.2832 |
| Indian Liver Patient | **83.2162±2.9526** | 83.1440±4.1555 | 83.1612±3.9406 | 83.1260±4.3989 |
| abalone | **51.8046±6.5298** | 22.7847±5.2576 | 32.1086±4.5430 | 49.3723±5.0002 |

and uses entropy regularization to optimize physical expressions, therefore, our algorithm cannot achieve satisfactory results in these two dataset. It can be viewed on the above analysis that PUFC has better classification performance when the number of labeled sample sets is extremely missing.

In addition, to gain a better understanding of our work, we add the number of labeled instances in the experimental process gradually. For instance, in Banknote authentication dataset, we tried a series of initial positive instances from 20% to 60%. Fig. 2 illustrates the experimental results with different number of labeled instances. It can be observed in the figure that the classification effect of the algorithm higher than comparison algorithms as a whole. However, the F-measure of PUFC algorithm has not been effectively improved due to the weak attribute correlation of dataset Wholesale customers or Indian Liver Patient. Another interesting observation from Fig. 2 is that SPY 、 Basic and Pruning are incapable to learn a suitable classifier, when the dataset Banknote authentication 、 Haberman Survival and abalone having few labeled instances for training. Its poor performance could be attributed to two main reasons: the great imbalance in the training sets (20% or 30% labeled instances against 70% unlabeled instances), and the difficulty of capturing the diversity in content and style of dataset from a small number of instances. Look at dataset abalone again, when there are more labeled instances, the comparison algorithms show more effectual improvement rate. While PUFC can obtain more comprehensive data distribution information even when the amount of labeled instances is very small, and then the classification effect is relatively stable when labeled instances are slightly increased. when the set of labeled instances are sufficient, the comparison algorithms also have strong expressiveness.

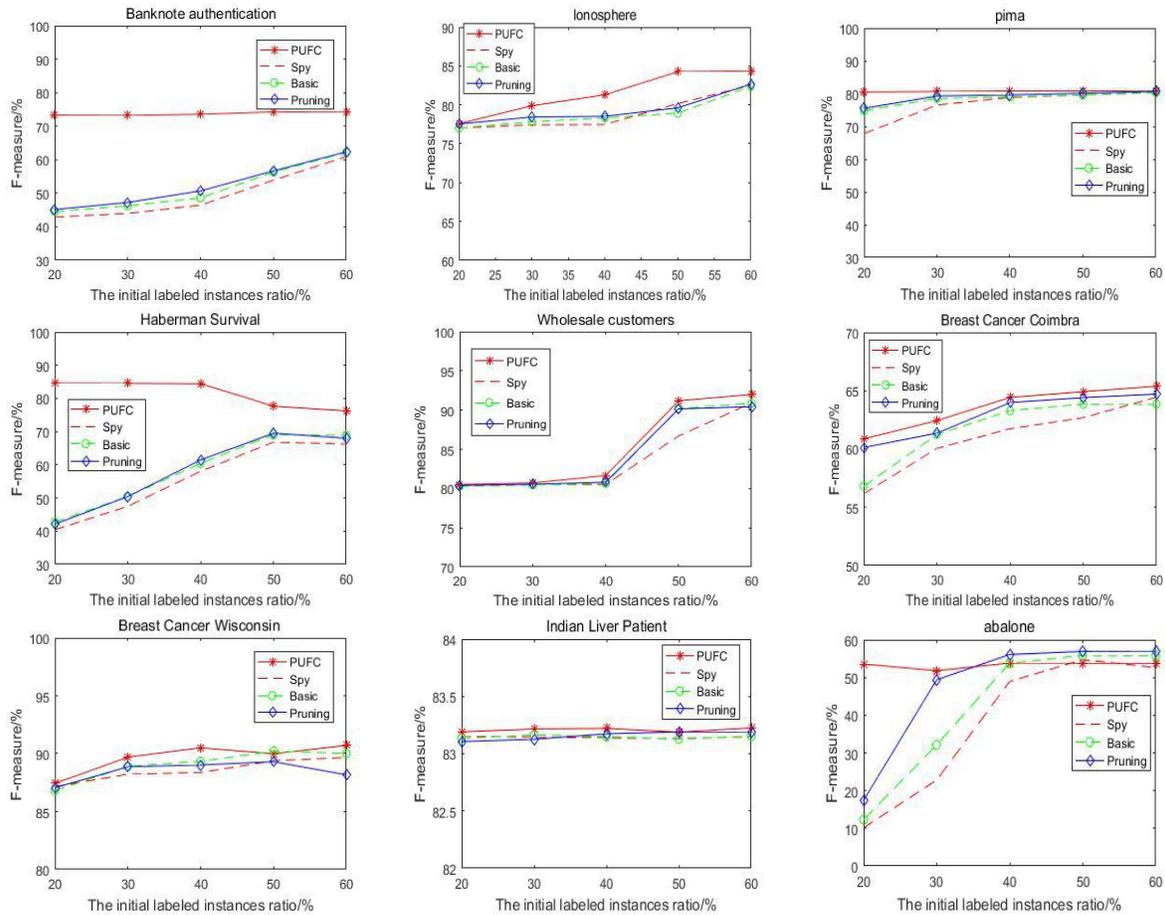

**Figure.2. The relationship between the average classification accuracy of four algorithms on nine datasets**

## 5 Conclusions

In reality, the number of labeled data is often small, and there are a huge number of unlabeled data. In this paper, the problem of drawing lessons from positive and unlabeled instances is tackled by using a novel approach called PUFC. PUFC could achieve good performance when the labeled instances proportion in training set was low. As future work we aim at applying the novel PU-learning to cope with the problem of data multi-classification.

## ACKNOWLEDGMENTS

This work is supported by Chongqing Normal University 2018 research and innovation programme, under award number YKC.18025.